# Active Sites model for the B-Matrix Approach

Krishna Chaithanya Lingashetty

*Abstract : This paper continues on the work of the B-Matrix approach in hebbian learning proposed by Dr. Kak. It reports the results on methods of improving the memory retrieval capacity of the hebbian neural network which implements the B-Matrix approach. Previously, the approach to retrieving the memories from the network was to clamp all the individual neurons separately and verify the integrity of these memories. Here we present a network with the capability to identify the "active sites" in the network during the training phase and use these "active sites" to generate the memories retrieved from these neurons. Three methods are proposed for obtaining the update order of the network from the proximity matrix when multiple neurons are to be clamped. We then present a comparison between the new methods to the classical case and also among the methods themselves.*

**Introduction**

Studying the human brain has always had scientists wondering as to the unique functioning of the organ[1]-[10]. How does the brain perceive all the information that it receives? How does the brain remember those particular memories? How does the brain then retrieve these memories when needed? Are the memories indexed? If so, then, by what criteria are they indexed? These are questions to which we cannot give a definite answer at this point in time. These questions are what form the motivation for the intense study being conducted on neural networks.

We can assume that in the millions of neurons, there is a hierarchy that exists among them and that specific parts of this hierarchy are to carry out several different physiological functions. In such a scenario, it is safe to assume that neural activity might start at critical points and then spread through the network until equilibrium is reached. This was the idea proposed by Dr. Kak in the B-Matrix approach [9],[10]. The B-Matrix approach provides a way of retrieving memories from the network by activity spreading. Any external stimuli to the network, for example, an electrical signal from any other organ, would spread through the network based on the geometric proximity of the neurons to each other. This signal would reach neighboring neurons first and then to the others as specified by the proximity matrix, a matrix which defines how the neurons are arranged in 3-dimensional space, of the neural network.

In the B-Matrix approach, the lower triangular matrix of the weight matrix is used for reconstruction of memories. The input signal is considered as a fragment of memory and this fragment keeps updating as the activity spreads through the network. This updated fragment is then fedback to the network for further retrieval and this fragment grows by following the update order derived from the proximity matrix. The activity spread from one neuron to another is calculated iteratively, as follows, until the fragment of memory becomes the complete memory.

$$f_i = sgn(B_i\, f_{i-1})$$

$$where\ f_i\ is\ the\ fragment\ of\ length\ i,$$

$$B_i\ is\ the\ i^{th}\ row\ of\ the\ B-Matrix$$

$$and\ sgn\ is\ the\ signum\ function.$$

$$T = B + B^T$$

The assumption is that this signal that is given to a particular neural network would be directed at a particular set of neurons, which give us a better chance at retrieving memories from the network. a very different approach is presented in [13],[14] for indexing. In this paper, we define "active sites" in the neural network that are identified at training state, which would serve as the sites at which the neural network would be stimulated with different stimuli. This poses the question of how the activity would spread from these activation sites, as there might be more than one way in which activity might spread from multiple neurons. Three methods are proposed in this paper, which would address activity spread in the network.

**Active Sites**

The motivation behind defining active sites, comes from the idea that different organs in the human body generate signals which are then directed to specific parts of the brain. Here we assume that this particular signal is not only sent to a specific neural network, but also a particular neuron in the neural network, the idea being, to target the right neuron with the right stimulus, so as to generate the specified memories that are stored in the network. A set of activations sites would then retrieve to us a memory, and a different set might retrieve another.

Consider a network consisting of 'n' neurons. According to the Hebbian rules of learning, the activations/synaptic strengths that are contained within the network, n(n-1)/2 in number, can be represented in an 'n x n' matrix with the lower and upper triangles containing the specified strengths between successive neurons. By saying that a network can identify a neuron as a possible active site, we propose that the network not only assigns a synaptic strength between two neurons, but also assigns an activation level for the specified neuron. In such a case, when the network is stimulated externally, it is these activation sites that are triggered and these are the sites from where activity spread takes place.

**Identifying Active sites**

The neural network identifies the active sites in itself during training. Since the assumption is that "triggering the right neurons with the right stimulus, yields a successful memory retrieval", identifying the right neurons forms the crux of this problem. For a given set of memories, say, M [1 : m] having a hamming distance of at least 1 among each other and a neural network of size n, we try to identify those sites in M[i] which are distinct from the memories of M[1 : m] − M[i]. Indirectly, we are trying to find those neurons for a particular memory, M[i], where the input for that particular neuron is different than the other M[1 : m]-M[i] memories.

After identifying the various active sites for a particular memory, the neural network then assigns an activation level on these neurons. Hence, different memories have different activation sites. The network cannot identify which activation belongs to which memory, but it can differentiate between activation levels of neurons corresponding to different memories based on the level of activation. For the simulation purposes, the activation levels are represented by prime numbers.

Now that we have established how active sites are identified, we now move onto the problem of trying to retrieve memories from these active sites. Here, we are posed with various questions such as "how many of these active sites do we use to retrieve memories?", "how do we generate update orders for potentially non-successive active sites?". For this, we propose three methods for determining an update order.

**Determining the Update Order**

Previously, as we were clamping single neurons, the update order of a neuron was as dictated by the corresponding row of the proximity matrix for that particular neuron. However, while considering active sites, there is always a possibility that there are more than one active site for a particular memory, and that the activity spread should ideally take place from these sites until the network reaches a stable state. In order to generate such an order, we propose three methods

- Arbitrary : Here, the neurons in the network are selected arbitrarily from the set of neurons for the update order.
- Averaged : Here we consider the average of the proximity matrix with regard to the synaptic strength between the active sites and then obtain the new update order relative to these active sites.
- Independent : Here, the partial memories retrieved from all active sites are taken individually along with the values obtained from the (B x clamp), these (B x clamp) values are then taken as a summation over the set of active sites to determine the resultant memory retrieved from the network.

**Decreasing the Computational Complexity**

Consider the case where a neural network of size 64 is used and the number of memories that are fed to the network is 6. Assume that the length of the fragment that is being considered for retrieval from this given neural network is 4. In such a case, we need to consider the number of combinations in which the sites are chosen in the classical B-Matrix approach. The total number of combinations of fragments and their originators are characterized by the selection of the number of sites and the fragment with which it is being stimulated.

$$O(BM) = \sum(combination\ of\ neurons * possible\ fragments)$$

$$O(BM) = (64c_1 * 2^1) + (64c_2 * 2^2) + (64c_3 * 2^3) + (64c_4 * 2^4)$$

$$O(BM) = \sum_{i=1}^{r} nc_i * 2^i$$

*where n is the size of the network*

*and r is the chosen fragment size*

Consider that the active sites model is presently being used. In such a scenario, the number of neurons that are presently under consideration are the neurons which are chosen as active sites for each memory. Hence, every memory has at most 'r' such sites from which we need to retrieve memories.

$$O(AS) = \sum (combination\ of\ neurons * possible\ fragments)$$

$$O(AS) = (4c_1 * 2^1) + (4c_2 * 2^2) + (4c_3 * 2^3) + (4c_4 * 2^4)$$

$$O(AS) = \sum_{i=1}^{r} rc_i * 2^i$$

*where r is the chosen fragment size*

As we can notice, the number of executions of the retrieval process are reduced drastically, which would help reduce the computational complexity of the network.

**Results**

The following graphs show the results obtained from the three methods that were discussed above. Table 1 depicts how successful memory retrieval is as the size of the network increases using the arbitrary method.

| Number of Neurons | Trained Memories | Successful Retrieval |
|---|---|---|
| 12 | 8 | 3.4 |
| 16 | 8 | 3.2 |
| 20 | 8 | 3.2 |
| 24 | 8 | 3 |

Table 1

| Number of Neurons | Trained Memories | Successful Retrieval |
|---|---|---|
| 12 | 8 | 2.4 |
| 16 | 8 | 2.4 |
| 20 | 8 | 2.3 |
| 24 | 8 | 2.1 |

Table 2

| Number of Neurons | Trained Memories | Successful Retrieval |
|---|---|---|
| 12 | 8 | 4.5 |
| 16 | 8 | 4.3 |
| 20 | 8 | 4 |
| 24 | 8 | 3.8 |

Table 3

The tables above show that the memory retrieval from the neural network is high, when the size of the neural network is comparatively smaller. As the size of the network grows, the memory retrieval capacity of the network is decreasing. Although, the pattern that the three methods follow seem to be similar, the rate of the decrease in the memory retrieval capacity does seem to increase across these methods.

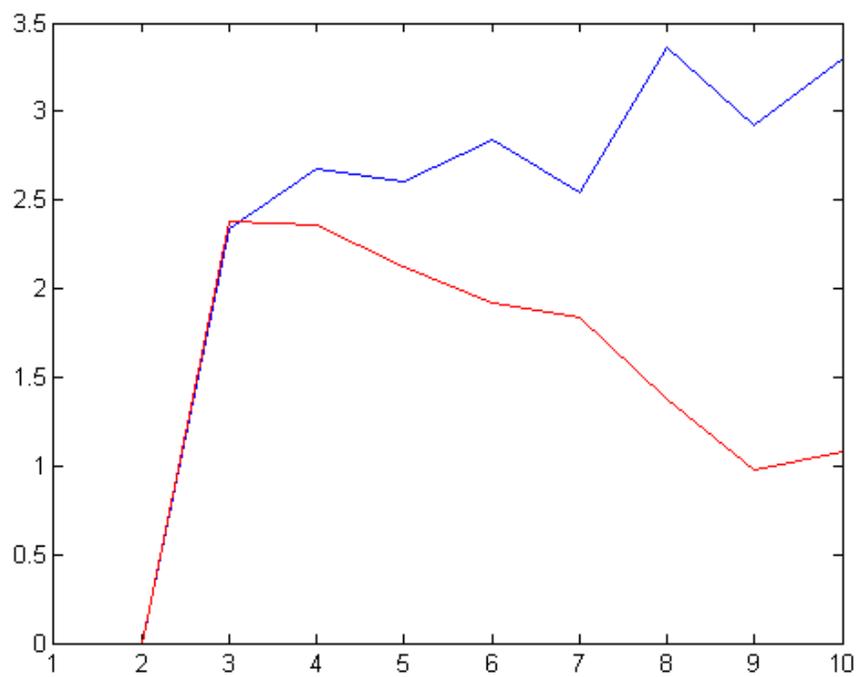

Figure 1 : Arbitrary, 16 neurons, 250 iterations

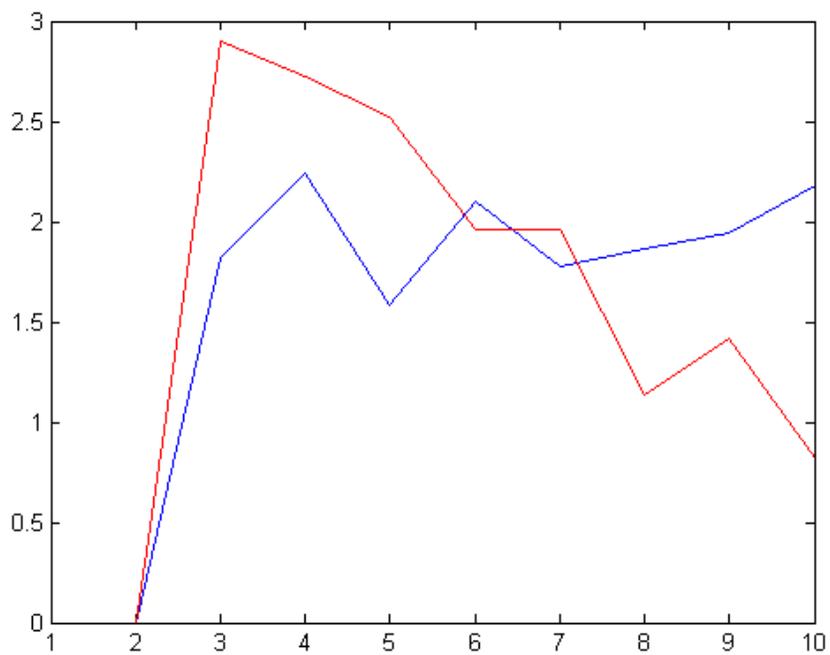

Figure 2 : Averaged, 24 neurons, 250 iterations.

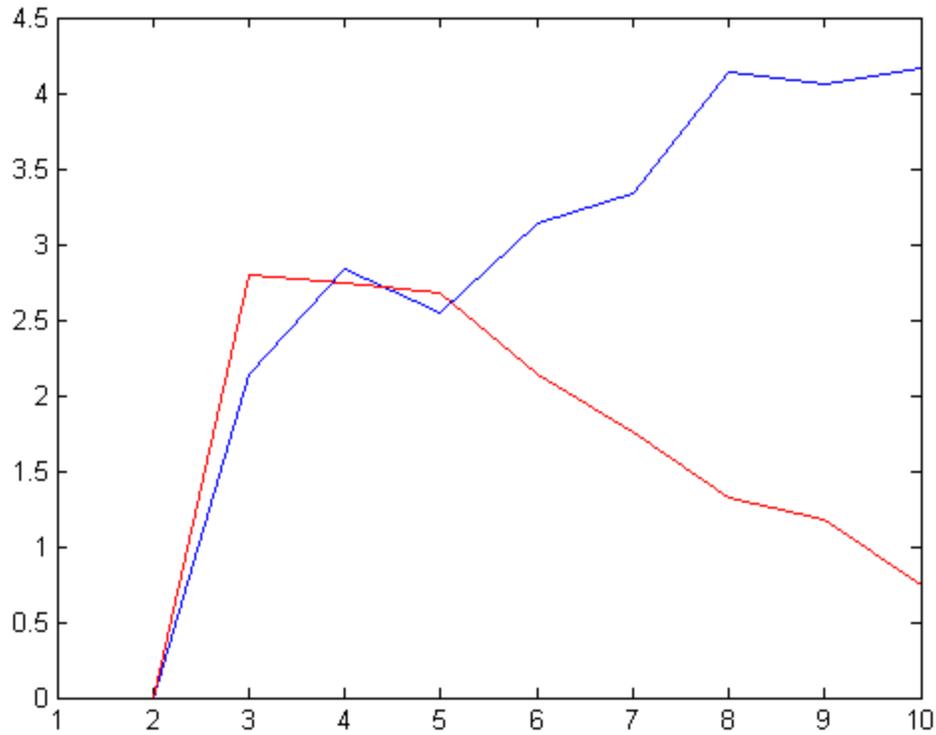

Figure 3 : Independent, 16 neurons, 250 iterations.

From the above figures, we can note that the number of memories that are being retrieved are comparably more than the number of memories that were retrieved from the previous approaches. We also notice that the Averaged method does not give us a better chance at retrieving memories when compared to the Arbitrary or the Independent methods. We also notice that as the number of memories being trained to the network increases, the number of verified memories gradually decreases, but the methods proposed do show an increase in successful retrieval.

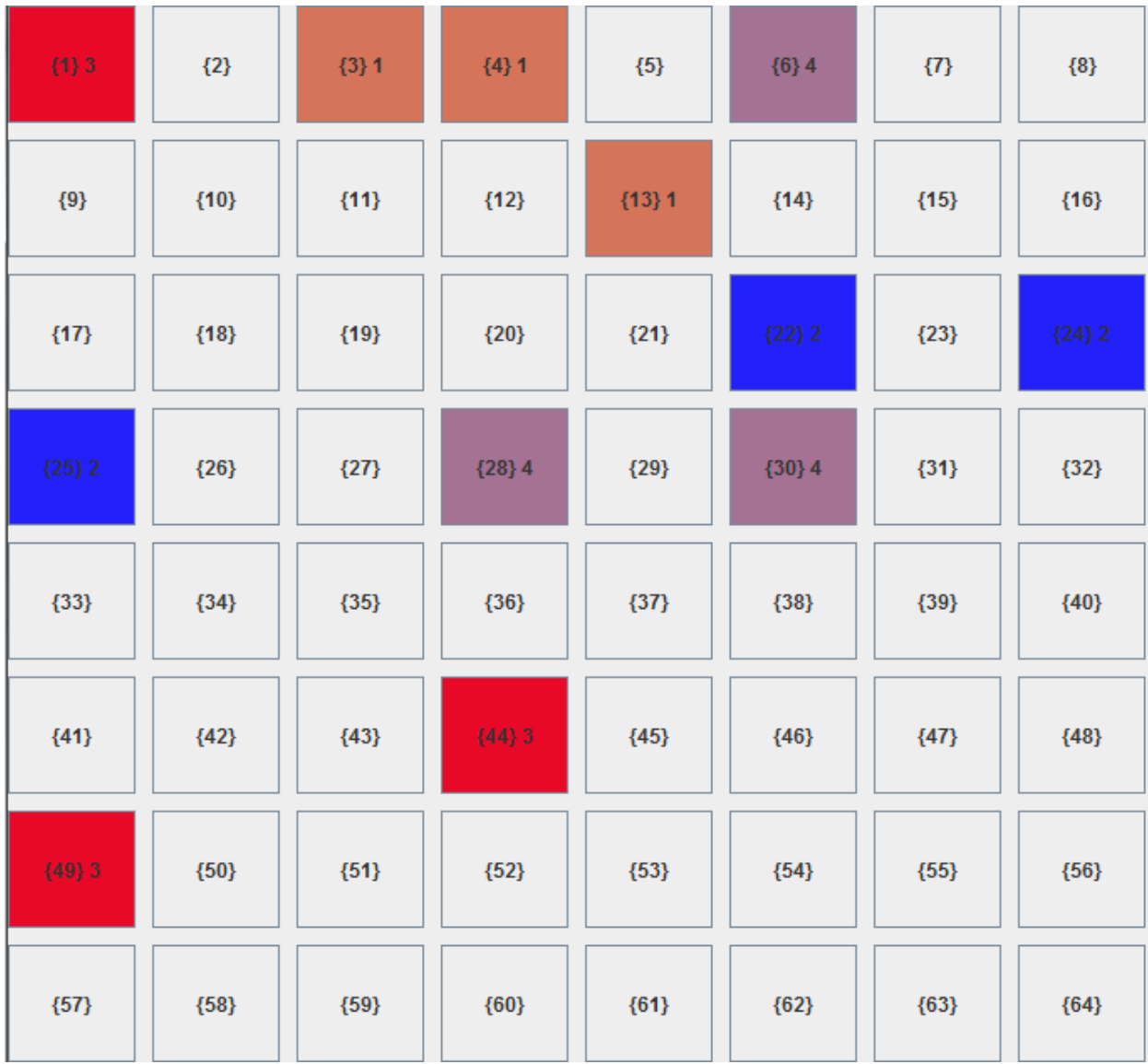

The above graph shows the 2-dimensional spread of the active sites through the neural network. each color represents a different set of active sites pertaining to a separate memory. These groups of neurons give us a better chance at memory retrieval than the classical approach which reduces the computational complexity.

**Conclusion**

In conclusion, we have been able to sustain a retrieval capacity even though the number of memories being fed to the neural network have been increasing. This can be attributed to the fact that as we keep accumulating new memories, some of our old memories fade away. The proposed model thus, incorporates the B-Matrix approach and determines a way of storing or retrieving memories with better computational complexity and maintaining a sustained retrieval of memories. It also provides a different approach to how the human brain might actually function.

**References**

[1] S. Zeki, A Vision of the Brain, Blackwell Scientific Publications, Oxford, 1993.
[2] D.O. Hebb, The Organization of Behavior. Wiley, 1949.
[3] S. Kak, The three languages of the brain: quantum, reorganizational, and associative. In: K. Pribram, J. King (Eds.), Learning as Self- Organization, Lawrence Erlbaum, London, 1996, pp. 185-219.
[4] S. Kak, The honey bee dance language controversy. Mankind Quarterly 31, 357-365, 1991.
[5] M.F. Bear, B. Connors, and M. Paradiso, Neuroscience: Exploring the Brain. Lippincott Williams and Wilkins, 2006.
[6] S. Kak, Better web searches and prediction with instantaneously trained neural networks, IEEE Intell. Syst. 14 (6), 78–81, 1999.
[7] S. Kak, Can we define levels of artificial intelligence? Journal of Intelligent Systems, vol. 6, pp.133-144, 1996.
[8] J.J. Hopfield, Neural networks and physical systems with emergent collective computational properties. Proc. Nat. Acad. Sci. (USA), vol. 79, pp. 2554-2558, 1982.
[9] S. Kak, Single neuron memories and the network‟s proximity matrix. 2009. arXiv:0906.0798
[10] S. Kak, Feedback neural networks: new characteristics and a generalization. Circuits, Systems, and Signal Processing, vol. 12, pp. 263-278, 1993.
[11] S. Kak, Self-indexing of neural memories. Physics Letters A, vol. 143, pp. 293-296, 1990.
[12] S. Kak, State generators and complex neural memories. Pramana, vol. 38, pp. 271-278, 1992.
[13] S. Kak and M.C. Stinson, A bicameral neural network where information can be indexed. Electronics Letters, vol. 25, pp. 203-205, 1989.
[14] M.C. Stinson and S. Kak, Bicameral neural computing. Lecture Notes in Computing and Control, vol. 130, pp. 85-96, 1989.
[15] D.L. Prados and S. Kak, Neural network capacity using the delta rule. Electronics Letters, vol. 25, pp. 197-199, 1989.
[16] D. Prados and S. Kak, Non-binary neural networks. Lecture Notes in Computing and Control, vol. 130, pp. 97-104, 1989.
[17] S. Kak and J. F. Pastor, Neural networks and methods for training neural networks. US Patent 5,426,721, 1995.
[18] R.Q. Quiroga, L. Reddy, G. Kreiman, C. Koch, and I. Fried, Invariant visual representation by single neurons in the human brain. Nature 435, pp. 1102–1107, 2005.
[19] W.M. Kistler and W. Gerstner, Stable Propagation of Activity Pulses in Populations of Spiking Neurons. Neural Computation, vol. 14, pp. 987- 997, 2002.
[20] K.H. Pribram and J. L. King (eds.), Learning as Self-Organization. Mahwah, N. J.: L. Erlbaum Associates, 1996.
[21] K Grill-Spector, R. Henson, and A. Martin, Repetition and the brain: neural models of stimulusspecific effects. Trends in Cognitive Sciences, 2006.